# Compliant motion control for handling a single object by two similar industrial robots

Hanene Mkaouar[a], Olfa Boubaker[a]*

[a]*National Institute of Applied Sciences and Technology, Centre Urbain Nord, BP 676 - 1080 Tunis, Tunisia*

**Abstract**

In this paper, we propose a compliant motion control strategy for handling a single object by two similar industrial robots. The dynamics of the object carried by the two robots is assimilated to the dynamics of a mass-spring-damper system described by a piecewise linear model (PWA). The coordination of the two robots is accomplished using a master slave synchronization approach dedicated for PWA systems, based on the Lyapunov theory, and solved via Linear Matrix Inequalities (LMIs). The performances of the proposed approach are proved by simulation results and compared to a related approach.

## 1. Introduction

Several industrial tasks, demanding precision, require robot coordination. A specified operation consists of handling a single object [1]. Compliant motion tasks are manipulation tasks that involve contacts between the manipulated object and the robots in which the trajectory of the manipulators are modified depending on the contact forces [2]. Position/force control is usually used for solving such cooperation problems [3, 4, 5, 6, 7, 8]. In fact, the control in position can ensure a tracking of the desired trajectory of the constraint robot and the control in force ensures a specific desired behavior of the robot when it enters in contact with the handled object. Master slave synchronization can be considered as a promising solution for such problems [9, 10].

This paper proposes a master slave synchronization approach for solving a coordination problem of two similar robots handling a single object. The dynamics of the object carried by the two robots is assimilated to the dynamics of a mass-spring-damper system described by a piecewise linear model (PWA). The master slave synchronization controller is based on the Lyapunov theory and solved via Linear Matrix Inequalities (LMIs).

The paper will be organized as follows: The cooperation problem is presented in the second section. The third section presents the analogical spring-dumper-mass system. A master slave synchronization approach is then applied. The effectiveness of the proposed approach is shown by simulation results and compared to a related approach.

* Corresponding author. Tel.: +216-98-950869.
*E-mail address:* hanene.mkaouar@gmail.com

## 2. Problem formulation

We consider in this paper the cooperation problem of two industrial robots handling a single object as shown by Fig. 1. The cooperation problem consists on handling a single object and displacing it through a compliance strategy. The control problem will be solved as a master slave synchronization problem of positions and velocities of the two robots as shown by Fig.2.

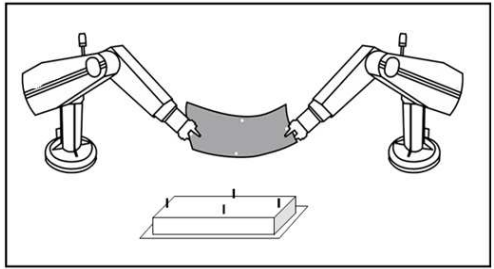
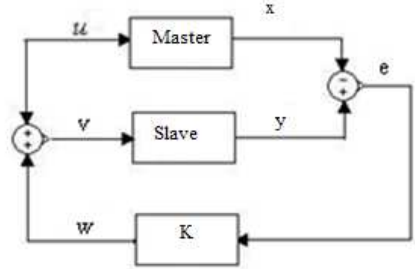

Fig. 1. Two similar robots handling a single object

Fig. 2. Controlled master-slave system

## 3. System modeling

To develop the master slave synchronization approach, an analogical system is chosen [9]. The analogue model shown by Fig.3 is a mechanical system composed of two masses m1 and m2 joined by a spring-dumper combination with a stiffness parameter k and a dumping coefficient c. Masses, springs and dumpers ensure programmable compliances between the robots and the environment (case of the damper c1), the robot and the object to be displaced (case of the dumper c and the spring k), or the object to be displaced and the second robot (case of the spring k2). The result is an elastic virtual behaviour that we want to give to the whole moving system thanks to the control law in order to prevent robots and object damaging. We assume, here, that the master robot is actuated by a variable force $u(t) = A_d \sin(wt)$ and the slave's one by the control force $v(t)$ to be computed.

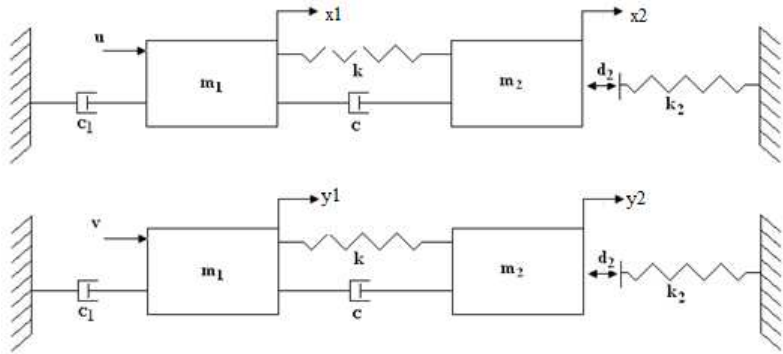

Fig. 3. Analogical spring-dumper-mass master and slave systems.

Using the Newton Euler formalism, the master dynamics are described by:

$$\begin{cases} \ddot{x}_1 = \dfrac{1}{m_1}[-k(x_1 - x_2) - c(\dot{x}_1 - \dot{x}_2) - c_1\dot{x}_1 + u] \\ \ddot{x}_2 = \dfrac{1}{m_2}[-k(x_1 - x_2) + c(\dot{x}_1 - \dot{x}_2) - k_2 g(x_2)] \end{cases} \quad (1)$$

whereas the slave dynamics are described by:

$$\begin{cases} \ddot{y}_1 = \dfrac{1}{m_1}[-k(y_1 - y_2) - c(\dot{y}_1 - \dot{y}_2) - c_1\dot{y}_1 + v] \\ \ddot{y}_2 = \dfrac{1}{m_2}[-k(y_1 - y_2) + c(\dot{y}_1 - \dot{y}_2) - k_2 g(y_2)] \end{cases} \quad (2)$$

where g ($x_2$) and g ($y_2$) are piecewise linear functions described, respectively, by:

$$\begin{cases} g(x_2) = x_2 - d_2 & \text{if } x_2 > d_2 \\ g(x_2) = 0 & \text{if } x_2 < d_2 \end{cases} \quad (3)$$

$$\begin{cases} g(y_2) = y_2 - d_2 & \text{if } y_2 > d_2 \\ g(y_2) = 0 & \text{if } y_2 < d_2 \end{cases} \quad (4)$$

The master and the slave systems can be, respectively, written in a piecewise linear form [11, 12, 13] as:

$$\dot{x} = A_j x + Bu + b_j \quad j = 1,2 \quad (5)$$

$$\dot{y} = A_i y + Bv + b_i \quad i = 1,2 \quad (6)$$

where:

$$A_1 = \begin{bmatrix} 0 & 1 & 0 & 0 \\ -\dfrac{k}{m_1} & -\dfrac{c+c_1}{m_1} & \dfrac{k}{m_1} & \dfrac{c}{m_1} \\ 0 & 0 & 0 & 1 \\ \dfrac{k}{m_2} & \dfrac{c}{m_2} & -\dfrac{k}{m_2} & -\dfrac{c}{m_2} \end{bmatrix}, A_2 = \begin{bmatrix} 0 & 1 & 0 & 0 \\ -\dfrac{k}{m_1} & -\dfrac{c+c_1}{m_1} & \dfrac{k}{m_1} & \dfrac{c}{m_1} \\ 0 & 0 & 0 & 1 \\ \dfrac{k}{m_2} & \dfrac{c}{m_2} & -\dfrac{k+k_2}{m_2} & -\dfrac{c}{m_2} \end{bmatrix}, b_1 = \begin{bmatrix} 0 \\ 0 \\ 0 \\ 0 \end{bmatrix}, b_2 = \begin{bmatrix} 0 \\ 0 \\ 0 \\ \dfrac{k_2 d_2}{m_2} \end{bmatrix}, B = \begin{bmatrix} 0 \\ \dfrac{1}{m_1} \\ 0 \\ 0 \end{bmatrix}$$

$x = [x_1 \ x_2 \ x_3 \ x_4]$ and $y = [y_1 \ y_2 \ y_3 \ y_4]$, ($x_1, x_2$) and ($y_1, y_2$) are respectively master and slave positions of mass $m_1$ and $m_2$ and $x_3 = \dot{x}_1$, $x_4 = \dot{x}_2$, $y_3 = \dot{y}_1$ and $y_4 = \dot{y}_2$ are the velocities.

## 4. Motion synchronization

Consider the master slave PWA system described by:

$$\begin{cases} \dot{x} = A_j x + b_j + Bu \\ \dot{y} = A_i x + b_i + Bv \\ w = K(y - x) \end{cases} \quad (7)$$

where $A_i \in \Re^{n \times n}$, $A_j \in \Re^{n \times n}$, $b_i \in \Re^n$ $b_j \in \Re^n$ are two constant matrices and two constant vectors, respectively. $B \in \Re^{n \times m}$ is the control matrix. $K$ is a gain state matrix of appropriate dimension.

Denote by $\Lambda_j$ and $\Lambda_i$ the partition of the state-space into polyhedral cells defined respectively by the following polytopic description:

$$\Lambda_j = \{x \mid H_j^T x + h_j < 0\} \quad (8)$$

$$\Lambda_i = \{z \mid H_i^T x + h_i < 0\} \quad (9)$$

where $H_j \in \Re^{n \times r_j}$, $h_j \in \Re^{r_j \times 1}$, $H_i \in \Re^{n \times r_i}$ and $h_j \in \Re^{r_i \times 1}$.

The error dynamics between the master and the slave systems can be written as:

$$\dot{e} = (A_i + BK)e + A_{ij} x + b_{ij} \quad (10)$$

where $e = y - x$, $A_{ij} = A_i - A_j$, $b_{ij} = b_i - b_j$.

*4.1. Theorem* [14]:

For a given decay $\alpha_1 > 0$ and for all pairs of indices $i, j \in Q$, if there exist constant symmetric positive definite matrix $S \in \Re^{n \times n}$, constant matrix $R \in \Re^{m \times n}$, diagonal negative definite matrices $E_{ij} \in \Re^{r_i \times r_i}$ and $F_{ij} \in \Re^{r_j \times r_j}$ and strictly negative constants $\beta_{ij}$ and $\xi_{ij}$, such that the following LMIs:

$$\begin{bmatrix} \xi_{ij} & \xi_{ij}|h_i|^T & \xi_{ij}|h_j|^T \\ * & \frac{1}{2}E_{ij} & 0 \\ * & * & \frac{1}{2}F_{ij} \end{bmatrix} < 0 \tag{11}$$

$$\begin{bmatrix} \Delta_{1,i} & A_{ij} & SH_i & 0 & \xi_{ij}b_{ij}|h_i|^T - \frac{1}{2}SH_iM_i & \xi_{ij}b_{ij}|h_j|^T \\ * & \beta_{ij}I & H_i & H_j & -\frac{1}{2}H_iM_i & -\frac{1}{2}H_jM_j \\ * & * & 2E_{ij} & 0 & 0 & 0 \\ * & * & * & 2F_{ij} & 0 & 0 \\ * & * & * & * & \frac{1}{2}E_{ij} - \xi_{ij}|h_i|\|h_i\|^T & -\xi_{ij}|h_i|\|h_j\|^T \\ * & * & * & * & * & \frac{1}{2}F_{ij} - \xi_{ij}|h_j|\|h_j\|^T \end{bmatrix} < 0 \tag{12}$$

$$\Delta_{1,i} = A_i S + S A_i^T + BR + R^T B^T + \alpha_1 S - \xi_{ij} b_{ij} b_{ij}^T$$

are satisfied, then the master-slave synchronization error system (10) is globally asymptotically stable using the state gain matrix:

$$K = RS^{-1} \tag{13}$$

*Proof*: see [14] and for more details [15].

*4.2. Simulation results*

Simulation results are conducted for the master and the slave systems (5) and (6) using the following parameters: $m_1 = 100$, $m_2 = 1$, $c = c_2 = 2$, $k = k_2 = 10$, $d_2 = 0.01$, $A_d = 1.5$ and $w = 1.5$ and for the initial conditions $x_0 = [1 \ 0.01 \ 0.01 \ 0.01]^T$ and $y_0 = [0.05 \ 0 \ 0.01 \ 1]^T$. The polyhedral cells: $\Lambda_1 = \{x|x_3 \geq d_2\}$ and $\Lambda_2 = \{x|x_3 < d_2\}$ are satisfying the polytopic description (8) and (9) where $H_1 = [0 \ 0 \ 1 \ 0]^T$, $H_2 = [0 \ 0 \ -1 \ 0]^T$, $h_1 = d_2$ and $h_2 = -d_2$.

Solving the LMIs (11) and (12) using the LMI toolbox of MatLab software for the parameter $\alpha_1 = 10^{-4}$ gives the state gain matrix (13) such as:

$$K = 10^3 [-1.7003 \ -0.4002 \ 0.7830 \ -0.2389]$$

Simulation results shown by Fig.4, where the synchronization of the positions ($x_1$, $y_1$) and ($x_2$, $y_2$) and the velocities ($\dot{x}_1$, $\dot{y}_1$) and ($\dot{x}_2$, $\dot{y}_2$) of the masses $m_1$ and $m_2$ are shown, prove the efficiency of the proposed approach. On the other hand, the smoothness of the compliant control law is shown by Fig.5.

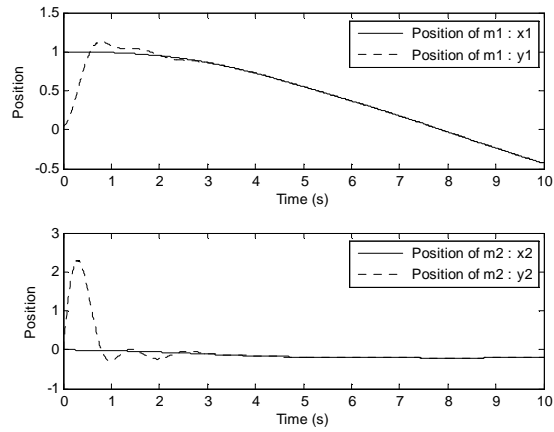

(a)

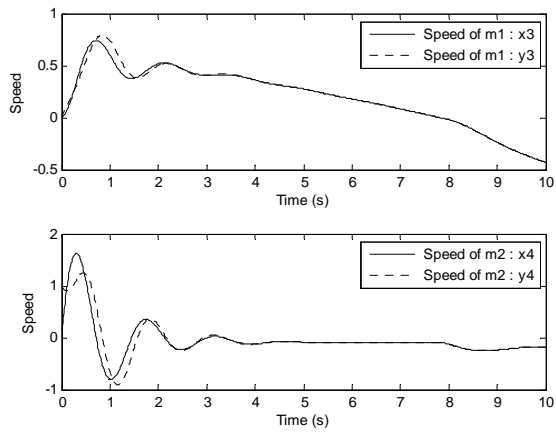

(b)

Fig. 4. Simulation results: (a) position synchronization of masses $m_1$ and $m_2$, (b) velocity synchronization of masses $m_1$ and $m_2$.

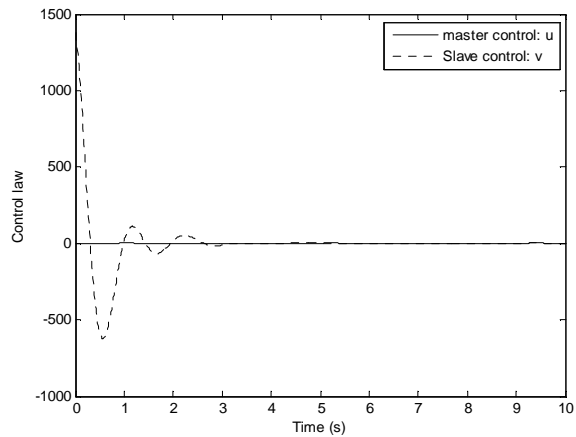

Fig.5. Compliant control law

*4.3. Comparative study*

To derive a comparative study, the synchronization problem of the master and the slave systems (5) and (6) is achieved using the strategy proposed in [16] using the same parameters. The computed state gain vector is given by:

$$K = [-35.2260 \ -6.5654 \ -12.1954 \ -9.8635]$$

The synchronization errors relating the two approaches are shown by Fig.6. It is clear that the two approaches give comparable results. In fact both approaches allow asymptotic synchronization but even if error variance is larger using approach [15] for the case study specified in this paper, tighter stability criteria are guaranteed by the approach [16], as it was already proved in [14], since it based on Lyapunov theory.

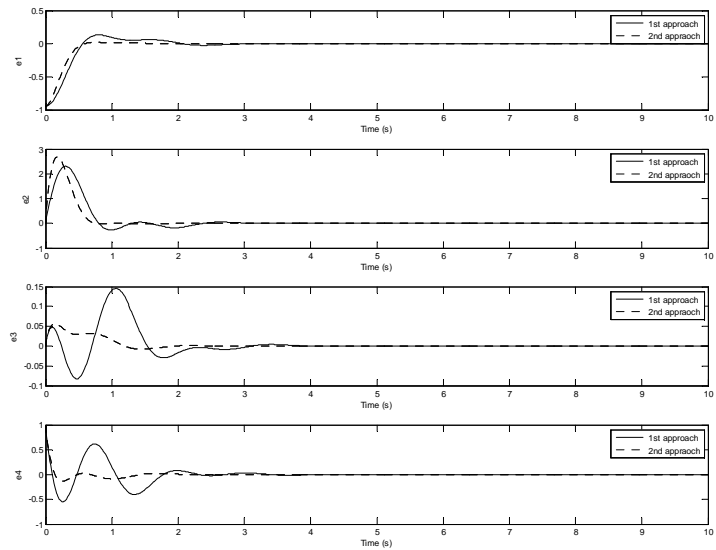

Fig. 6. Synchronization error using first approach [15] and second approach [16]

## 5. Conclusion

In this paper, a compliant motion controller is proposed for handling a single object by two similar industrial robots. The coordination was accomplished using an analogue piecewise linear dynamical system and a master slave synchronization approach. The efficiency of the proposed approach was proved via simulation results and finally compared to a related approach.